\documentclass[journal]{IEEEtran}

\usepackage{graphicx}
\usepackage{bm}
\usepackage{subfigure}
\usepackage{algorithm,algorithmic}
\usepackage{amsmath}
\usepackage{multirow}
\usepackage{multicol}
\usepackage{cite}
\usepackage{threeparttable}
\usepackage{color}
\usepackage{doi}
\usepackage{enumerate}
\usepackage{caption}
\usepackage{rotating}
\usepackage{multirow}
\usepackage{threeparttable}
\usepackage{booktabs}
\usepackage{makecell}
\usepackage{setspace}
\usepackage{amsmath}
\usepackage[square, comma, sort&compress, numbers]{natbib}

\begin{document}
\title{Multi-Objective Evolutionary Multi-tasking with Two-stage Adaptive Knowledge Transfer based on Population Distribution}
\author{Zhengping Liang, Weiqi Liang, Xiuju Xu, Xiaoliang Ma, Ling Liu, and Zexuan~Zhu,~\IEEEmembership{Member,~IEEE}
\thanks{This work has been submitted to the IEEE for possible publication. Copyright may be transferred without notice, after which this version may no longer be accessible. \par
Z. Liang, W. Liang, X. Xu, X. Ma, L. Liu and Z. Zhu are with the College of Computer Science and Software Engineering, Shenzhen University, Shenzhen 518060, China (e-mail: liangzp@szu.edu.cn; liangwq0131@foxmail.com; 377611694@qq.com; maxiaoliang@yeah.net; liulingcs@szu.edu.cn; zhuzx@szu.edu.cn).}}

\markboth{}
{Shell \MakeLowercase{\textit{et al.}}: Bare Demo of IEEEtran.cls for IEEE Journals}
\maketitle
\begin{abstract}
Multi-tasking optimization can usually achieve better performance than traditional single-tasking optimization through knowledge transfer between tasks. However, the current multi-tasking optimization algorithms suffer from some deficiencies. On the problems sharing high similarity, the knowledge that can accelerate the convergence rate of tasks has not been fully taken advantages of. On the less similar problems, the algorithms tend to suffer from negative transfer, which may result in performance degradation. In addition, some knowledge transfer methods proposed previously do not fully consider how to deal with the situation in which the population falls into local optima. To solve these issues, this paper proposes an evolutionary multi-tasking optimization algorithm for multi/many-objective optimization with two-stage adaptive knowledge transfer based on population distribution . The resultant algorithm namely EMT-PD can accelerate and improve the convergence performance of tasks based on the knowledge extracted from the probability model that reflects the search trend of the whole population. At the first transfer stage, an adaptive weight is used to adjust the step size of individual's search, which can reduce the impact of negative transfer. At the second stage of knowledge transfer, the individual's search range is further adjusted dynamically, which can improve the diversity of population and be beneficial for jumping out of local optima. Experimental results on multi-tasking multi-objective optimization test suites show that EMT-PD is superior to other six state-of-the-art evolutionary multi/single-tasking algorithms. To further investigate the effectiveness of EMT-PD on many-objective optimization problems, a multi-tasking many-objective optimization test suite is also designed in this paper. The experimental results on the new test suite also demonstrate the competitiveness of EMT-PD.
\end{abstract}
\IEEEpeerreviewmaketitle	

\section{Introduction}	
\IEEEPARstart {M}{ulti-objective} optimization problems (MOPs) widely exist in real world \cite{MOPs-1,MOPs-2,MOPs-3}. Generally, an MOP can be described as follows:
\begin{equation}
\begin{split}\label{e1}
&{\min_{\bm{x}} (F(\bm{x})=\left(f_{1}(\bm{x}), \ldots, f_{n}(\bm{x})\right))}
\\& {\text { subject to } \bm{x} \in \Omega^D},	
\end{split}
\end{equation}
where $\bm{x} = (x_1,...,x_D)$ represents a $D$-dimensional decision vector in search space $\Omega^D$. $F(\bm{x}) = (f_{1}(\bm{x}), \ldots, f_{n}(\bm{x}))$ denotes the objective function vectors with $n$ objectives. An MOP is also referred to as a many-objective optimization problem (MaOP) if $n>3$ \cite{NSGA-III,MaOPs-1,MaOPs-2}. A solution $\bm{x}$ is said to dominate another solution $\bm{y}$, if and only if $x_i\leq y_i$ for all $i\in [1,n]$ and there exists one objective $j\in [1,n]$ such that $x_j<y_j$. A solution not dominated by any other solutions is called a Pareto optimal solution. All Pareto optimal solutions form the Pareto optimal set (PS) of which the corresponding projection in the objective space is called the Pareto front (PF).

MOPs and MaOPs can be effectively solved by multi/many-objective evolutionary algorithms \cite{msmc_chen,msmc_tian,AR-MOEA,Ind-1,Ind-2,Ind-3,MOEA/D,Dec-1,DEC-2,DEC-3,NSGA-II,Dom-1,Dom-2,Dom-3}. The majority of the existing multi/many-objective evolutionary algorithms are designed to handle one problem at a time. To solve a new problem, the algorithms must start from scratch. However, many optimization problems encountered in real-world are correlated with each other. The experience of solving one problem can be beneficial to the solving of another related problem. In this line and inspired by the capability of human brain to process transactions in parallel, Gupta \textit{et al.} \cite{MFEA} proposed a new optimization paradigm namely evolutionary multi-tasking (EMT) to deal with multiple optimization tasks simultaneously. Compared to the single-tasking counterpart algorithms, EMT has been shown to be able to considerably improve the overall performance in solving multiple related optimization tasks via knowledge transfer.

Following \cite{MFEA}, a number of multi-objective EMT algorithms have been proposed in the literature to solve various MOPs. For example, Gupta \textit{et al.} extended \cite{MFEA} and proposed a multi-objective EMT algorithm (MOMFEA) \cite{MOMFEA} to deal with multiple MOPs simultaneously through assortative mating and vertical cultural transmission. Yang \textit{et al.} \cite{TMO-MFEA} presented a two-stage assortative mating method for EMT with decision variables classification, i.e., assortative mating is carried out on different variable groups with different parameters to balance the diversity and convergence. Feng \textit{et al.} \cite{EMT-EGT} proposed an EMT algorithm with explicit genetic transfer (EMT-EGT) to enhance the search ability of the evolution population. Chen \textit{et al.} \cite{MM-DE} introduced a memetic EMT framework for knowledge transfer between subpopulations. Tuan \textit{et al.} \cite{EMT-PSM} put forward an EMT algorithm employing local search strategy to accelerate the convergence of population. EMT has also achieved successes in real-world applications, e.g., permutation-based combinatorial optimization problems \cite{Application2}, branch testing problems in software engineering \cite{Application3}, modular knowledge representation in neural networks \cite{Application1}, symbolic regression problems \cite{GP-MFEA}, multi-objective pollution-routing problems \cite{Application5}, and hyperspectral image unmixing \cite{Application4}.

The research on EMT algorithms has made remarkably progress, yet there remain room for further improvement and open questions to be addressed. Particularly, on high similar problems, the knowledge of high quality solutions has not bee fully used to improve the convergence of the population. On problems with low similarity, the population distributions of the tasks are generally different, which tends to result in negative transfer between tasks \cite{EMT-VIEW}. Moreover, the existing knowledge transfer methods also do not take good care of the situation where the population falls into local optima.

To address the aforementioned issues, this paper proposes a new multi-objective EMT algorithm namely EMT-PD with two-stage adaptive knowledge transfer based on population distribution. EMT-PD firstly builds probability models for each task, and then obtains knowledge from the product of different probability models. The knowledge can help to accelerate the convergence rate of population. At the first stage of knowledge transfer, the step size of individual's search is adjusted by adaptive weight, which can reduce the probability of generating negative transfer. At the second stage of knowledge transfer, the search range of individual is further adjusted dynamically, which can promote the diversity of population and avoid getting trapped in local optima. EMT-PD is tested on multi-tasking multi/many-objective optimization test suites. The comparison studies with six state-of-the-art evolutionary multi/single-tasking algorithms demonstrate the competitiveness of EMT-PD. The main contributions of this paper are highlighted as follows:

1) A multi-tasking multi/many-objective evolutionary optimization algorithm with novel method of extracting and transferring knowledge is proposed to improve the efficiency and performance of optimization.

2) A multi-tasking many-objective optimization test suite is designed based on a representative many-objectives test suite MaF \cite{MaF}.

3) Based on three test suites, EMT-PD is fully analyzed by comparing with state-of-the-art algorithms.

The rest of this paper is organized as follows. Section II introduces the related work and motivation of EMT-PD. Section III describes the details of EMT-PD. Section IV presents the experimental design. Section V demonstrates the performance of EMT-PD with different probability models. In the end, Section VI concludes this work and discusses some potential future directions.

\section{Related Work and Motivation}
This section briefly reviews the related knowledge extract and transfer methods and explains the motivation of the proposed method.
\subsection{Extract and Transfer Knowledge in EMT}
In EMT algorithms, knowledge can be extracted from a single or multiple individuals and transferred to other individuals to facilitate their search. Particularly, knowledge transfer methods based on single individual (KTS) refer to extracting knowledge from a single individual of one task, and transferring knowledge to other tasks. EMT algorithms with KTS include MFEA \cite{MFEA}, M-BLEA \cite{M-BLEA}, LDA-MFEA \cite{LDA-MFEA}, S\&M-MFEA \cite{SM-MFEA}, MO-MFEA \cite{MOMFEA}, GMFEA \cite{GMFEA}, TMO-MFEA \cite{TMO-MFEA}, MTO-DRA \cite{MTO-DRA}, MFEA-II \cite{MFEA-II}, MFEA-GHS \cite{MFEA-GHS}, and MFGP \cite{msmc_zhong}. All the above algorithms transfer knowledge through assortative mating and vertical cultural transmission \cite{EMT-boon}. In assortative mating, two individuals are selected from the population randomly, and then generate offspring by simulated binary crossover (SBX) and polynomial mutation. In vertical cultural transmission, each offspring is randomly assigned to a task. In KTS, each individual provides different knowledge for the tasks, and the diversity of population is maintained effectively. However, EMT algorithms with KTS cannot fully utilize the knowledge of high quality solutions to accelerate the convergence rate of the population due to the randomness of knowledge transfer.

Knowledge transfer methods based on multiple individuals (KTM) refer to extracting knowledge from multiple individuals of one task, and transferring knowledge to other tasks. EMT algorithms with KTM can be implemented based on particle swarm optimization (PSO). For example, Feng \textit{et al.} \cite{MFPSO} proposed a multi-tasking PSO algorithm. The convergence of population is accelerated by the guidance of optimal solutions of multiple tasks. Tang \textit{et al.} \cite{AMFPSO} proposed an adaptive multi-tasking PSO algorithm by introducing a self-adaption strategy to adjust the inter-task knowledge transfer probability, which reduces the probability of negative transfer effectively. Song \textit{et al.} \cite{MTMSO} proposed a multi-tasking multi-swarm optimization algorithm. The quality of the solutions is improved by crossover between optimal individuals of all tasks.

Differential evolution (DE) is another popular platform to construct EMT with KTM. For instance, Liu \textit{et al.} \cite{EMT-DE-1} proposed SaM-MA, with three different mechanisms employed, i.e., DE algorithm, predicting optimal solution via surrogate model, and local searching strategy. The mixed use of these three mechanisms can balance diversity and convergence of population. Zhou \textit{et al.} \cite{EMT-DE-2} proposed a new mutation strategy called DE/best/1$+\rho$, with gradually increased weight of knowledge transfer in the process of evolution, which improves the diversity of population.

Explicit knowledge transfer represents another form of KTM in EMT algorithms. Feng \textit{et al.} \cite{EMT-EGT} first proposed an EMT algorithm EMT-EGT with explicit knowledge transfer, which allows the incorporation of multiple search mechanisms with different biases. It is able to improve the search ability of population. Shang \textit{et al.} \cite{EMT-EGT-1} proposed a credit assignment approach, which selects proper individuals for explicit knowledge transfer. The efficiency of knowledge transfer of this method was shown to be improved.

KTM is beneficial to improve the convergence of population, but it also has the probability of extracting knowledge from inferior individuals. Besides, the search of entire population at one generation is guided by the same solutions in some KTMs, which increase the probability of population falling into local optima.

\begin{figure}
\captionsetup{font={footnotesize}}
\centering
\fontsize{8}{9}\selectfont
\captionsetup{font={small}}
\subfigure[]{
\begin{minipage}{4cm}
\includegraphics[width=4cm,height=3.7cm]{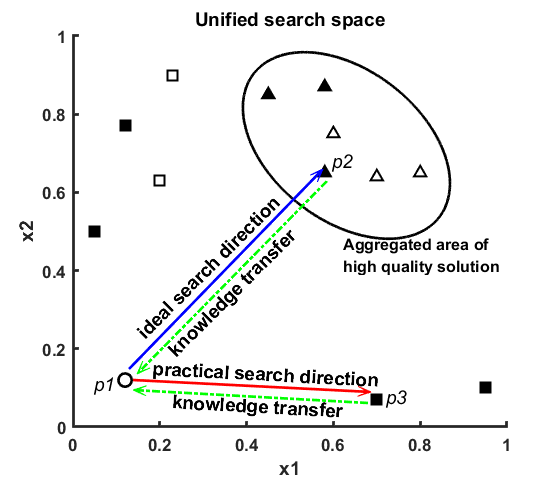}
\end{minipage}%
\label{F1a}
}%
\subfigure[]{
\begin{minipage}{4cm}
\includegraphics[width=4cm,height=3.7cm]{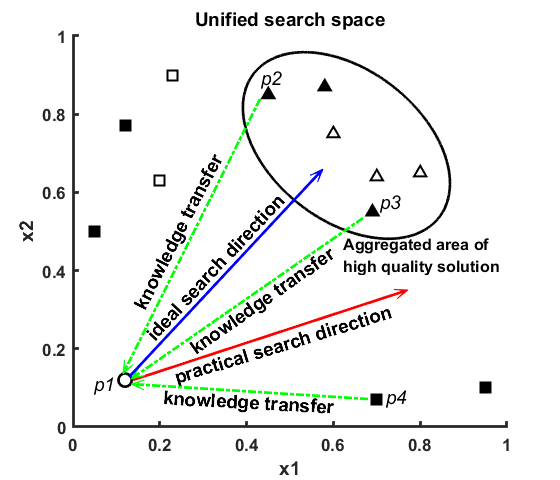}
\end{minipage}%
\label{F1b}
}%
\\
\subfigure[]{
\begin{minipage}{4cm}
\includegraphics[width=4cm,height=3.7cm]{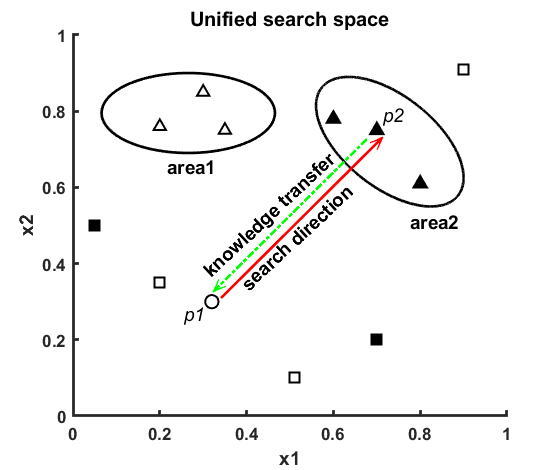}
\end{minipage}%
\label{F1c}
}%
\subfigure[]{
\begin{minipage}{4cm}
\includegraphics[width=4cm,height=3.7cm]{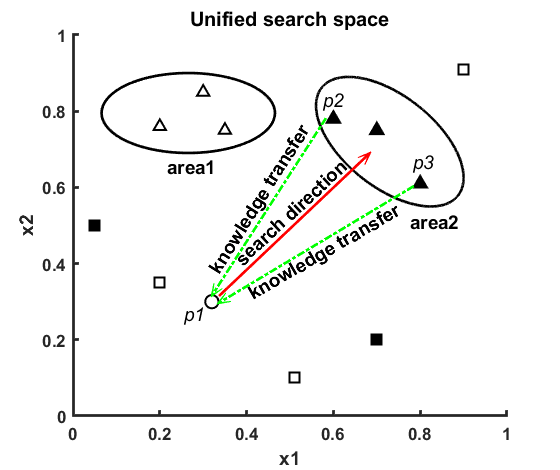}
\end{minipage}%
\label{F1d}
}%

\caption{The illustration of knowledge transfer and search direction, where x1 and x2 are the two dimensions of the decision variables. Triangles and squares represent the high quality solutions and inferior solutions, respectively. The solutions of task1 and task2 are distinguished with hollow and solid faces, respectively. $p1$ represents an solution selected from task1. The global optima are supposed to be located in the ellipse areas. (a) and (b) are KTS and KTM in high similarity scenario, respectively. (c) and (d) are KTS and KTM in low similarity scenario, respectively.}
\label{F1}
\end{figure}	

\subsection{Motivation of Two-stage Adaptive Knowledge Transfer based on Population Distribution}
In this subsection, we take an example of two-tasking problem as shown in Fig. 1 to explain the motivation of the proposed with two-stage adaptive knowledge transfer based on population distribution.

For high similarity problems, knowledge extracted from the high quality solutions of one task can accelerate the convergence of another task effectively \cite{GMFEA}. However, for both KTS and KTM, knowledge may be extracted from inferior individuals. Fig. 1\subref{F1a} shows an example of KTS in high similarity scenario. In the search space, high quality solutions of task1 and task2 converge in the same area, $p1$ is a solution of task1, $p2$ is a high quality solution of task2, and $p3$ is an inferior solution of task2. If knowledge is transferred from $p2$ to $p1$, then $p1$ will be guided to search the centralized area of high quality solutions. However, KTS may transfer knowledge from inferior solution $p3$ to $p1$, which would slow down the convergence of task1. Fig. 1\subref{F1b} shows an example of KTM in high similarity scenario where a solution is guided by two other solutions. $p1$ is a solution of task1. $p2$ and $p3$ are two high quality solutions of task2, respectively. $p4$ is a inferior solution of task2. If knowledge is transferred from $p2$ and $p3$ to $p1$, $p1$ will search the centralized area of high quality solutions. However, KTM may also transfer the knowledge of inferior solution $p4$ to $p1$, if $p1$ learns from $p3$ and $p4$, which would cause $p1$ to deviate from the ideal search direction.

\begin{algorithm}[h]
  \caption{Main Framework of EMT-PD}
  \label{A1}
  \begin{algorithmic}[1]
    \REQUIRE
      $R$, type of probability model.
    \ENSURE
      a series of non-dominated solutions.
    \STATE
      Initialize population $\bm{P}$;
    \STATE
      Split $\bm{P}$ into two subpopulations $\bm{pop}_{1}$ and $\bm{pop}_{2}$;
    \WHILE {\emph{stopping conditions are not satisfied}}
      \STATE
        Build probability model for $\bm{pop}_{1}$ and $\bm{pop}_{2}$ by \textbf{Algorithm  \ref{A2}};
      \STATE
        Calculate the maximum points of product of two probability models \boldmath $\bm{mp} $ \unboldmath according to Eq. (\ref{E5});
      \STATE
        Conduct two-stage adaptive knowledge transfer and generating offspring $\bm{C}$ by \textbf{Algorithm \ref{A3}};
      \STATE
        Evaluate offspring;
      \STATE
        Environmental selection;
    \ENDWHILE
\end{algorithmic}
\end{algorithm}

For low similarity problems, the population distribution of task1 and task2 is very different \cite{EMT-synergy}. Whether in KTS or KTM, population of different tasks are difficult to guide each other to search with ideal search direction. There would be a high probability of generating negative transfer between those tasks. Fig. 1\subref{F1c} shows an example of KTS in low similarity scenario, high quality solutions of different tasks are distributed in different areas. $p1$ represents an individual of task1. $p2$ is a high quality solution of task2. When knowledge is transferred from $p2$ to $p1$, $p1$ will deviate severely from the convergence area of high quality solutions of task1. Fig. 1\subref{F1d} shows an example of KTM in low similarity scenario, which is similar to the situation of KTS. Moreover, in KTMs implemented with PSO or DE, the search of the entire population at one generation is only guided by the best solutions of the tasks, which easily leads the population to local optima.

To address the above issues, this paper presents a novel EMT algorithm with two-stage adaptive knowledge transfer based on population distribution. Specifically, the probability models are firstly built for the population of each task, and the knowledge used for transfer is extracted from the maximum point of the probability models' product. Note that the population distribution rather than individual(s) is used as the knowledge source to relieve the impact of inferior individuals. Afterward, at the first stage of knowledge transfer, the step size of individual's search is adaptively adjusted to reduce the impact of negative transfer. At the second stage of knowledge transfer, the search range of individual is adjusted again dynamically, which can increase the diversity of population and reduce its probability of falling into local optimum.

\section{PROPOSED ALGORITHM}
Multi-objective optimization algorithms have achieved in many practical applications \cite{MOPs-1,MOPs-2,MOPs-3}. However, the potential synergies between distinct optimization problems has not been fully explored in traditional multi-objective optimization algorithms. In this section, we propose a multi-tasking algorithm with two-stage knowledge transfer base on population distribution to exploit the potential synergies between problems. In this section, we first introduce the main framework of EMT-PD. Then the process of building probability model and extracting knowledge is proposed. After that, the details of two-stage adaptive knowledge transfer are explained. Finally, computational complexity analysis of EMT-PD is discussed.
\subsection{The Main Framework of EMT-PD}
The main framework of EMT-PD is summarized in \textbf{Algorithm \ref{A1}}. Firstly, the population $\bm{P}$ is initialized, and divided into two sub-populations $\bm{pop}_{1}$ and $\bm{pop}_{2}$ for different tasks. At each evolutionary generation, a probability model is built for each task by \textbf{Algorithm \ref{A2}} in line 4. In line 5, the maximum point $\bm{mp}$ of the product of probability models is obtained by Eq. (\ref{E5}), which is used  to guide the search of population. The two-stage adaptive knowledge transfer and offspring generation are carried out by \textbf{Algorithm \ref{A3}} in line 6. It is worth noting that EMT-PD is a general algorithm, which supports various types of probability model.

\begin{algorithm}
  \caption{Build Probability Model}\label{A2}
  \begin{algorithmic}[1]
    \REQUIRE
      $pop$, the subpopulation of task1 or task2; $R$, the type of probability model.
    \ENSURE
      $\bm{M}$, the probabilistic model; $\bm{m}$, maximum point of $\bm{M}$.
    \FOR {\emph{each dimension of decision variable}}
      \STATE
        Generate the log-likelihood $LL(\theta_{j})$ by Eq. (\ref{E2});
      \STATE
        Build probability model $M_{j}$ for $pop$ by Eq. (\ref{E3});
      \STATE
        Calculate the maximum point $m_{j}$ of $M_{j}$ by Eq. (\ref{E4});
    \ENDFOR
\end{algorithmic}
\end{algorithm}

\subsection{Build Probability Model}
In this paper, the Maximum Likelihood Estimation (MLE) is used to estimate the parameters of probability model according to population, and the maximum point of probability model is obtained.

Let $M$ denote the probability models of population. $\theta_{j}$ is the parameter of $M$ in $j$-$th$ dimension of decision variable. The MLE is used to find the parameter value $\hat{\theta}_j$, which maximizes the log-likelihood $LL(\theta_{j})$. Firstly, $LL(\theta_{j})$ is calculated as follows:

\begin{equation}\label{E2}
LL(\theta_{j})=\sum_{i=1}^{N}ln f(p_{i,j}|\theta_{j}),
\end{equation}
where $p_{i,j}$ represents the $j$-$th$ variance of $i$-$th$ individual, $i = 1, 2, ..., N$. $N$ denotes the population size. The type of probability model $f(p_{i,j}|\theta_{j})$ is $R$. $ln$ denotes the natural logarithm.

Then, $\hat{\theta_{j}}$ is calculated as follows:

\begin{equation}\label{E3}
\hat{\theta_{j}}=\underset{\theta_{j}}{\arg\max}LL(\theta_{j}).
\end{equation}

Based on above calculations, the probability model of $j$-$th$ decision variable $M_{j}$ be obtained, that is $f(p_{i,j}|\hat{\theta}_{j})$, which also can be labeled as follows:

\begin{equation}\label{E3A}
M_j(x)=f(x\mid \hat{\theta_{j}}).
\end{equation}

The maximum point $\bm{m}$ of probability model $\bm{M}$ learned from population reflects the centralization of population in each generation of evolution \cite{Multi-Source}. The formula of calculating $m_j$ is as follows:

\begin{equation}\label{E4}
m_{j}=\underset{x}{\arg\max}M_{j}(x).
\end{equation}

\textbf{Algorithm \ref{A2}} shows the process of building probability models by MLE. For each dimension of decision variable, the log-likelihood $LL(\theta_{j})$ is calculated by Eq. (\ref{E2}) in line 2. Then the parameters of probability model $M_{j}$ are calculated by Eq. (\ref{E3}) in line 3. After that, the maximum point $m_{j}$ of $M_{j}$ is obtained according to Eq. (\ref{E4}) in line 4.

\begin{figure}[h]
\centering
\fontsize{8}{9}\selectfont
\captionsetup{font={small}}
\includegraphics[width=5cm,height=4cm]{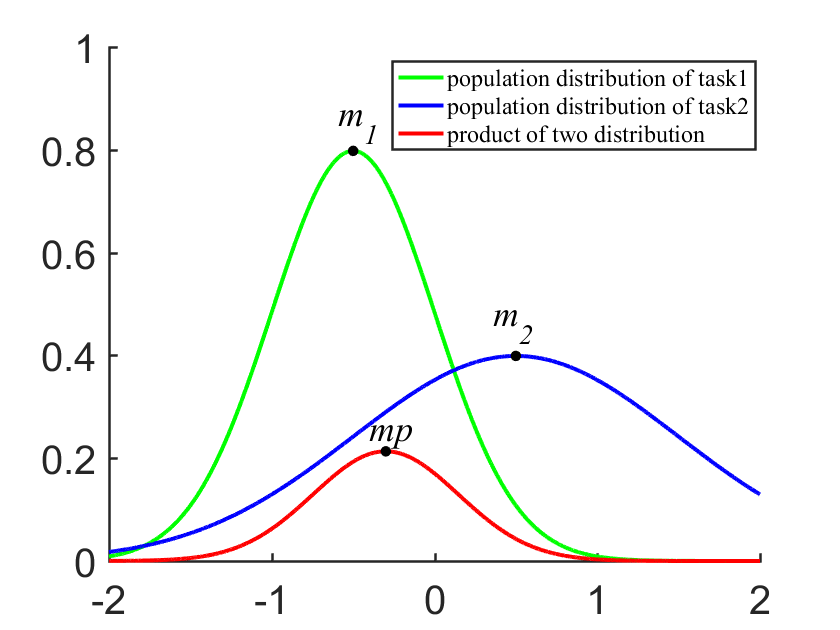}
\caption{The maximum point of probability models’ product reflects the common centralization of two populations.}
\label{F3}
\end{figure}

Finally, the probability model $\bm{M}$ and the maximum point $\bm{m}$ of $\bm{M}$ is obtained, where $\bm{M}=(M_1,M_2,...,M_j,...,M_D)$, $\bm{m}=(m_1,m_2,...,m_j,...m_D)$, and $D$ is the dimension of decision variable.

\subsection{Extract Knowledge from Population Distribution}
The maximum point of probability models’ product reflects the common centralization of two populations \cite{Multi-Source}. In our algorithm, the maximum point of probability models’ product is used to guide the search of each task. It is helpful to accelerate convergence rate and decrease the probability of falling into local optimum.

\begin{figure}
\center
\fontsize{8}{9}\selectfont
\captionsetup{font={small}}
\includegraphics[width=4.5cm,height=4cm]{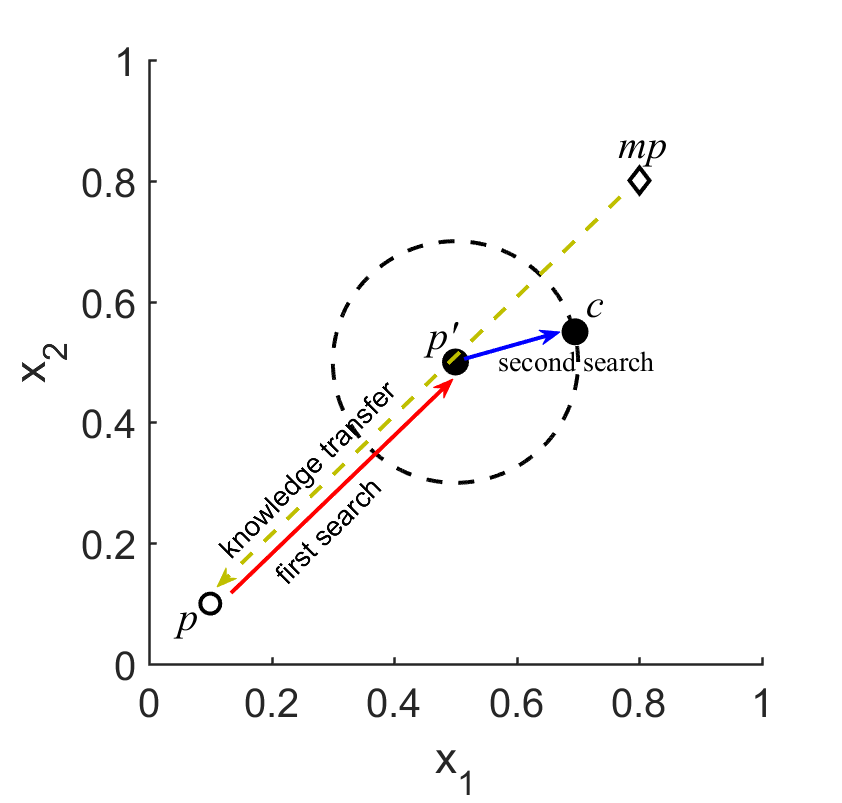}
\caption{An example of two-stage knowledge transfer in the 2-Dimensions unified searching space. x$_1$ and x$_2$ are the two dimensions of decision variable.}\label{F4}
\end{figure}

According to \textbf{Algorithm \ref{A2}}, the two probability models $\bm{M}_{1}$ and $\bm{M}_{2}$ can be learned from population of task1 and task2, respectively, where $\bm{M}_{1} = (M_{1,1},M_{1,2},...,M_{1,j},...,M_{1,D})$ and $\bm{M}_{2} = (M_{2,1},M_{2,2},...,M_{2,j},...,M_{2,D})$. The maximum point $\bm{mp}=(mp_1,mp_2,...,mp_j,...,mp_D)$ of the product of $\bm{M}_1$ and $\bm{M}_2$ can be calculated as follows:

\begin{equation}\label{E5}
mp_{j}=\underset{x}{\arg\max}(M_{1,j}(x)*M_{2,j}(x)).
\end{equation}

As shown in Fig. \ref{F3}. The green line is the population distribution of task1. The blue line is the population distribution of task2. $\bm{m}_1$ and $\bm{m}_2$ are the maximum point of task1 and task2 respectively. The red line represents the product of population distributions of task1 and task2, which reflects the common search trend of task1 and task2. The maximum point $\bm{mp}$ of red line reflects the common centralization of populations of task1 and task2, and $\bm{mp}$ is used to guide the search of each population.

\begin{algorithm}
  \caption{Two-stage Adaptive Knowledge Transfer and Offspring Generation}\label{A3}
  \begin{algorithmic}[1]
    \REQUIRE
      $\bm{pop}$, the subpopulation of task;
      $\bm{mp}$, the maximum point of product of two probability models;
      $\bm{m}$, the maximum point of the probability model of $\bm{pop}$.
    \ENSURE
      $\bm{C}$, offspring.
    \STATE
      Calculate the Euclidean distance $d_{1}$ of $\bm{m}$ and $\bm{mp}$ according to Eq. (\ref{E6});
    \FOR {\emph{each individual $p \in \bm{pop}$}}
      \STATE
        Calculate the Euclidean distance $d_{2}$ of $\bm{m}$ and $p$ according to Eq. (\ref{E7});
      \STATE
        Generate an intermediate individual $p'$ according to Eq. (\ref{E8});
      \STATE
        Generate an offspring $c$ according to $p'$ by Eq. (\ref{E10});
      \STATE
        $c'$ = Polynomial mutation($c$);
      \STATE
        $C = C\cup c'$;
    \ENDFOR
\end{algorithmic}
\end{algorithm}

\subsection{Two-stage Knowledge Transfer and Offspring Generation}
For high similarity problems, populations of task1 and task2 will converge to the approximate area, and the convergence rate of task1 and task2 can be accelerated by $\bm{mp}$ directly. But for low similarity problems, populations of task1 and task2 will converge to different areas. If $\bm{mp}$ is directly used to guide the search of population, there may be a lot of negative transfers. In addition, it is necessary to balance the convergence and the diversity of populations no matter for high or low similarity problems. To solve those issues, we propose a two-stage adaptive knowledge transfer based on population distribution. At the first stage, the step size of individual's search is adjusted adaptively to reduce the impact of negative transfer. At the second stage, the search range of individual is further based on an intermediate individual, increases the diversity of population and helps to jumping out of local optimum.

Fig. \ref{F4} displays the process of two-stage knowledge transfer, where $p$ is a individual selected from population of certain task. $\bm{mp}$ is the maximum point of product of probability models. At the first stage of knowledge transfer, the knowledge is transferred from $\bm{mp}$ to $p$. The search of $p$ is guided by $\bm{mp}$, and an intermediate individual $p'$ is generated by the first stage knowledge transfer. At the second stage, the search range of the intermediate individual $p'$ is adjusted again dynamically to generate offspring $c$.

\textbf{Algorithm \ref{A3}} is the pseudo code of the two-stage adaptive knowledge transfer and offspring generation for one task. Firstly, the Euclidean distance $d_{1}$ between $\bm{mp}$ and the maximum point $\bm{m}$ of the task's probability model is calculated in line 1. Then for each individual $p$, the Euclidean distance $d_{2}$ between $p$ and $\bm{m}$ is calculated in line 3. $d_{1}$ and $d_{2}$ are calculated as follows:

\begin{equation}\label{E6}
d_{1}=\sqrt{tr((\bm{m}-\bm{mp})^T(\bm{m}-\bm{mp}))},
\end{equation}

\begin{equation}\label{E7}
d_{2}=\sqrt{tr((\bm{m}-p)^T(\bm{m}-p)},
\end{equation}

where $tr()$ denotes the trace operation of a matrix.
In line 4, the intermediate individual $p'$ is generated via the first stage of knowledge transfer. $p'$ is calculated is as follows:

\begin{equation}\label{E8}
p'=p+w(\bm{mp}-p),
\end{equation}
where $w$ is the adaptive weight of knowledge transfer. $w$ is defined as follows:

\begin{equation}\label{E9}
w = d_{2}/(d_{1}+d_{2}).
\end{equation}

As $\bm{m}$ reflects the centralization of population which $p$ belongs to, if $\bm{mp}$ is close to $\bm{m}$, it means that the population distribution of the two tasks are similar, and the knowledge from $\bm{mp}$ can effectively guide the search of $p$. Therefore, it is suitable to increase the wight of knowledge transfer $w$ when $\bm{mp}$ is close to $\bm{m}$. In \textbf{Formula \ref{E9}}, $d_1$ is the distance between $\bm{mp}$ and $\bm{m}$, the smaller $d_1$ is, the larger $w$ is. On the contrary, if $\bm{m}$ is far away from $\bm{mp}$, the difference of the population distribution of two tasks will become large, and the knowledge from $\bm{mp}$ may leads negative transfer between tasks. It is necessary to decrease the wight $w$ of knowledge transfer in the situation. In \textbf{Formula \ref{E9}}, the larger $d_1$ is, the smaller $w$ is. In addition, $d_2$ is used to measure the distance between individual $p$ and the centralization of population. If $d_2$ is small, it means that $p$ is closer to the centralization of population. The weight $w$ needs to be decreased to reduce transfer distance of $p$. In \textbf{Formula \ref{E9}}, the smaller $d_2$ is, the smaller $w$ is. On the contrary, if $d_2$ is large, it means $p$ is far away from the centralization of population. The weight $w$ needs to be increased to enlarge transfer distance of $p$. In \textbf{Formula \ref{E9}}, $w$ increases with $d_2$.

In line 5, the offsping $c$ is generated via the second stage of knowledge transfer as follows:
\begin{equation}\label{E10}
c = p'+v,
\end{equation}
where $p'$ is an intermediate individual generated by first stage knowledge transfer, and $v$ is a search vector defined as follows:
\begin{equation}\label{E11}
v = \frac{1}{D}*F*Q*(d_{1}+d_{2}),
\end{equation}
where $D$ is the dimension of decision variable, $Q$ is the $D$-dimensional Gaussian White Noise, and $F$ is the scaling factor. The Gaussian white noise $Q$ adopted in this paper follows a standard Gaussian distribution, and its mean and variance are 0 and 1, respectively. Parameter $F$ is used to adjust the order of magnitude of white noise. White noise is introduced to increase population diversity.

The reason to introduce the second stage knowledge transfer is to better balance the convergence and the diversity of populations since all individuals are searched in the same direction guided by $\bm{mp}$ at the first stage. When $d_1$ is small, the knowledge transfer at the first stage may be effective due to the high similarity between the two tasks. It is not necessary to substantially adjust the range of searching at the second stage in Eq. (\ref{E11}). On the contrary, if d1 is large, it means the low similarity between two tasks. The adjustment of search range at the second stage should be reasonably increased. As for $d_2$, if it is small, the individual $p$ is close to the convergence area. It is also not necessary to substantially adjust again at the second stage. If $d_2$ is large, the adjustment of search range at the second stage also should be reasonably increased.
Polynomial mutation is performed in line 6. Offspring $c'$ is combined into population $\bm{C}$ in line 7.
\subsection{Computational Complexity Analysis}
In this section, the computational complexity of EMT-PD within a generation is discussed. EMT-PD mainly consists of three parts: 1) building probability model; 2) two-stage knowledge transfer and offspring generation; and 3) environmental selection. The maximum likelihood estimation is used to build probability model, which requires computational complexity of $O(DN)$. In two-stage knowledge transfer and offspring generation, a computational complexity of $O(DN)$ is needed. In environmental selection, computational complexity of non-dominated sorting and crowding distance are $O(nN^{2})$ and $O(nN\log(N))$, respectively. To sum up, the computational complexity of EMT-PD is $O(nN^{2})$, where $D$ and $N$ represent the dimension of decision variable and population size respectively, and $n$ is the number of objectives.

\section{EXPERIMENT AND ANALYSIS}
The performance of EMT-PD is firstly evaluated and compared with several state-of-the-art optimization algorithms on multi-tasking multi-objective test suites. Then a novel multi-tasking many-objective test suite is proposed, and the performance of EMT-PD on it is also evaluated and compared with state-of-the-art optimization algorithms. Due to space limitation, the experiment of convergence and running time are put into the Setion \uppercase\expandafter{\romannumeral3} and \uppercase\expandafter{\romannumeral4} of the \textbf{supplementary materials}.

\subsection{Experiment and Analysis on Multi-objective Problems}
\subsubsection {Test Problems and Compared Algorithms}
To assess the performance of EMT-PD, two multi-objective test suites are applied.

The classical multi-tasking multi-objective test suite MTMOPs \cite{MTMOPs}, which can be split into three groups according to the degree of intersection, i.e., complete intersection (CI), partial intersection (PI), and no intersection (NI). Moreover, these groups can be further partitioned into high similarity (HS), medium similarity (MS), and low similarity (LS). Therefore, there are nine problems, namely, CIHS, CIMS, CILS, PIHS, PIMS, PILS, NIHS, NIMS, and NILS.

The complex multi-tasking multi-objective optimization test suite CEC2019-CMO was proposed in IEEE CEC2019 competition on evolutionary multi-tasking optimization. CEC2019-CMO contains ten multi-tasking multi-objective problems. The details of CEC2019-CMO are summarized in \cite{CEC2019-CMO}.

Six state-of-the-art algorithms are used in comparision with EMT-PD, including three multi-objective EMT algorithms, i.e., MO-MFEA \cite{MOMFEA}, TMO-MFEA \cite{TMO-MFEA}, EMT-EGT \cite{EMT-EGT}, and three representative multi-objective evolutionary algorithms, i.e., NSGA-II \cite{NSGA-II}, AR-MOEA \cite{AR-MOEA} and CMOPSO \cite{CMOPSO}. NSGA-II is a basic multi-objective evolutionary algorithm serving as the baseline here. AR-MOEA is an indicator-based multi-objective optimization algorithms with an enhanced inverted generational distance indicator. CMOPSO is a multi-objective particle swarm optimization algorithm with competitive mechanism.

\subsubsection {Performance Metric}
The Inverted Generational Distance (IGD) \cite{IGD} is widely used to evaluate the performance of multi-objective algorithms. IGD is calculated as follows:

\begin{table}[htbp]
  \centering
  \linespread{1.2}
  \captionsetup{font={small}}
  \setlength{\tabcolsep}{3mm}{
  \fontsize{8}{9}\selectfont
  \begin{threeparttable}
  \caption{COMMON PARAMETER SETTINGS OF ALGORITHMS}\label{T1}
    \begin{tabular}{cc}
    \toprule
    Parameter&Value\cr
    \midrule
    \renewcommand{\arraystretch}{2}
    Size of population ($N$)&200\\
    Maximal iteration ($G$)&1000\\
    Maximal function evaluations ($EFs$)&200000\\
    Crossover probability ($p_{c}$)&0.3\\
    Mutation probability ($p_{m}$)&$1 \slash N$\\
    Distribution index for crossover ($\eta_c$)&20\\
    Distribution index for mutation ($\eta_m$)&20\\
    \bottomrule
    \end{tabular}
    \end{threeparttable}
    }
\end{table}

\begin{equation}\label{e12}
IGD({{P}^{*}}, \bm{A})=\frac{1}{\left| {{P}^{*}} \right|}\sum\limits_{\bm{z}\in {{P}^{*}}}{{\min  d(\bm{z},\bm{A})}},
\end{equation}
where $d(\bm{z},\bm{A})$ refers to the Euclidean distance between a reference point $\bm{z}$ and a solution $\bm{A}$ in the objective space, and $P^*$ represents a predefined set of reference points on the PF. The smaller IGD is, the better the convergence and diversity of population is.

\subsubsection {Parameter Settings}
In EMT-PD, the type of probability model $R$ is set to Gaussian probability model, and the scale factor $F$ is set to 0.01. The parameters of TMO-MFEA are set according to \cite{TMO-MFEA}, i.e., $rmp$ is set to 1 for the diversity variable and 0.3 for the convergence variable. According to the description of EMT-EGT in \cite{EMT-EGT}, SPEA2 is used for one task and NSGA-II is used for the other task. The interval of explicit transmission is set to 10. The size of the elite population $\lambda$ of CMOPSO is set to 10 according to \cite{CMOPSO}. Common parameter settings of algorithms are summarized in \textbf{Table \ref{T1}}.

It is worth noting that $F$ is set to a lower weight. There are two main reasons. The first is that the scale of decision variable space is between 0 and 1 after normalization. If a relatively high noise is chosen, it may cause the individual to exceed the search range of decision variable. The second is that the individual which experienced the first knowledge transfer has moved towards the centralized area of high quality solution. If a relatively large disturbance is introduced, it easily drives individual jumping out of the centralized area of high quality solutions, leading to the individual quality deterioration.  The sensitivity experiment results of parameter $F$ can refer to Section II of the \textbf{supplementary materials}.

\subsubsection {Results and Analysis on MTMOPs}
\textbf{Table \ref{T2}} shows the experimental results on MTMOPs. The best result is highlighted. The Wilcoxon rank sum test is performed at a significance level of 5$\%$. Symbols '+','$-$','$\approx$' denote that the result is significantly better, significantly worse, or comparable with that of EMT-PD, respectively. Next, the experimental results are analyzed from the degree of similarity.

\begin{table*}[htbp]
\centering
\linespread{1.2}
\captionsetup{font={small}}
\fontsize{8}{9}\selectfont
\caption{AVERAGED IGD VALUE OBTAINED BY EMT-PD, MO-MFEA, TMO-MFEA, EMT-EGT, NSGA-II, ARMOEA AND CMOPSO ON OVER 30 INDEPENDENT RUNS ON MTMOPs}\label{T2}	
\setlength{\tabcolsep}{3mm}{
\begin{tabular}{ccccccccc}
\toprule
Problem&Task&EMT-PD&MO-MFEA&TMO-MFEA&EMT-EGT&NSGA-II&ARMOEA&CMOPSO\\
\midrule
\multirow{2}{*}{CI$+$HS}&T1&\textbf{9.10E-04}&1.43E-02(-)&9.14E-03(-)&1.24E-03(-)&4.42E-01(-)&3.11E-01(-)&2.59E-01(-)\\
                        &T2&\textbf{9.66E-03}&3.22E-01(-)&6.58E-02(-)&6.62E+01(-)&3.53E-01(-)&3.83E-01(-)&2.34E-01(-)\\
\hline

\multirow{2}{*}{CI$+$MS}&T1&6.84E+00&7.34E+00(-)&1.64E+01(-)&5.34E+00(+)&8.47E+00(-)&\textbf{4.92E+00}(+)&7.17E+00(-)\\
                        &T2&\textbf{2.76E-03}&4.84E-03(-)&4.72E-03(-)&4.35E-03(-)&8.09E-03(-)&4.72E-03(-)&4.62E-03(-)\\
\hline

\multirow{2}{*}{CI$+$LS}&T1&\textbf{2.07E-03}&1.58E-02(-)&2.04E-02(-)&2.02E-02(-)&2.92E+01(-)&9.96E+00(-)&8.95E+00(-)\\
                        &T2&\textbf{9.54E-04}&6.59E-02(-)&2.04E-03(-)&2.21E+00(-)&2.23E-01(-)&2.26E-01(-)&2.23E-01(-)\\
\hline

\multirow{2}{*}{PI$+$HS}&T1&\textbf{1.46E-01}&1.49E-01(-)&1.29E-01(-)&2.43E-01(-)&2.01E-01(-)&1.94E-01(-)&2.25E-01(-)\\
                      &T2&1.30E+01&2.28E+01(-)&2.11E+01(-)&2.06E+01(-)&2.48E+01(-)&1.18E+01(+)&\textbf{1.18+01}(+)\\
\hline

\multirow{2}{*}{PI$+$MS}&T1&\textbf{1.33E-02}&3.39E-02(-)&2.27E-01(-)&2.98E-02(-)&4.41E-02(-)&1.39E-02(-)&1.39E-02(-)\\
                        &T2&\textbf{2.90E+01}&7.92E+02(-)&5.21E+02(-)&5.80E+02(-)&5.80E+02(-)&4.47E+02(-)&2.92E+02(-)\\
\hline

\multirow{2}{*}{PI$+$LS}&T1&1.57E-02&3.23E-02(-)&2.26E-02(-)&\textbf{5.31E-03}(+)&7.70E-02(-)&6.70E-03(+)&5.92E-03(+)\\
                        &T2&\textbf{1.73E+01}&3.81E+01(-)&2.59E+01(-)&2.09E+01(-)&5.77E+01(-)&1.73E+01($\approx$)&1.73E+01($\approx$)\\
\hline

\multirow{2}{*}{NI$+$HS}&T1&\textbf{5.09E-02}&4.94E+01(-)&5.00E+01(-)&4.73E+01(-)&3.56E+02(-)&8.80E+01(-)&1.48E+01(-)\\
                        &T2&\textbf{1.60E-03}&4.35E-02(-)&1.71E-02(-)&1.66E-02(-)&6.74E-01(-)&2.05E-01(-)&2.25E-01(-)\\
\hline

\multirow{2}{*}{NI$+$MS}&T1&\textbf{7.28E+00}&2.40E+01(-)&1.72E+01(-)&1.62E+01(-)&6.82E+01(-)&2.85E+01(-)&2.37E+01(-)\\
                        &T2&\textbf{9.89E-04}&1.17E+00(-)&8.16E-03(-)&7.75E-03(-)&4.05E+00(-)&3.28E+00(-)&2.70E+00(-)\\
\hline

\multirow{2}{*}{NI$+$LS}&T1&\textbf{6.27E-03}&6.53E-02(-)&2.07E-02(-)&2.87E-02(-)&6.66E-02(-)&1.10E-02(-)&1.09E-02(-)\\
                        &T2&\textbf{6.08E-02}&2.06E+01(-)&2.10E+01(-)&2.15E+01(-)&3.40E+01(-)&1.93E+01(-)&2.06E+01(-)\\
\hline
\multirow{1}{*}{$+/-/\approx$}&&&0/18/0&0/18/0&2/16/0&0/18/0&3/14/1&2/15/1\\
\bottomrule
\end{tabular}}
\end{table*}

\begin{table*}[htbp]
\centering
\linespread{1.2}
\captionsetup{font={small}}
\caption{AVERAGED IGD VALUE OBTAINED BY EMT-PD, MO-MFEA, TMO-MFEA, EMT-EGT, NSGA-II, ARMOEA AND CMOPSO OVER 30 INDEPENDENT RUNS ON CEC2019-CMO}\label{T3}	
\fontsize{8}{9}\selectfont
\setlength{\tabcolsep}{3mm}{
\begin{tabular}{ccccccccc}
\toprule
Problem&Task&EMT-PD&MO-MFEA&TMO-MFEA&EMT-EGT&NSGA-II&ARMOEA&CMOPSO\\
\midrule
\multirow{2}{*}{CMO1}&T1&3.78E-03&7.06E-03(-)&7.08E-03(-)&\textbf{3.72E-03}(+)&7.98E-03(-)&6.44E-03(-)&5.75E-03(-)\\
                             &T2&\textbf{3.12E-03}&3.49E-02(-)&3.34E-02(-)&3.51E-02(-)&4.01E-02(-)&5.61E-02(-)&1.72E-02(-)\\
\hline

\multirow{2}{*}{CMO2}&T1&\textbf{3.27E-03}&5.34E-03(-)&5.24E-03(-)&3.81E-03(-)&8.00E-03(-)&6.29E-03(-)&5.80E-03(-)\\
                             &T2&\textbf{5.51E-03}&1.40E-02(-)&1.36E-02(-)&2.93E-02(-)&1.91E-01(-)&2.52E-01(-)&6.49E-02(-)\\
\hline

\multirow{2}{*}{CMO3}&T1&1.25E-02&8.55E-02(-)&7.58E-02(-)&\textbf{1.02E-02}(+)&1.14E-01(-)&1.21E-01(-)&5.85E-02(-)\\
                             &T2&\textbf{7.77E-03}&2.90E-02(-)&2.71E-02(-)&1.82E-02(-)&4.05E-02(-)&4.50E-02(-)&2.30E-02(-)\\
\hline

\multirow{2}{*}{CMO4}&T1&\textbf{9.87E-03}&6.46E-02(-)&4.31E-02(-)&1.37E-02(-)&1.13E-01(-)&1.30E-01(-)&6.09E-02(-)\\
                             &T2&\textbf{1.04E-02}&6.60E-02(-)&4.64E-02(-)&1.79E-02(-)&1.44E-01(-)&1.48E-01(-)&5.53E-02(-)\\
\hline

\multirow{2}{*}{CMO5}&T1&8.78E-03&3.97E-02(-)&3.97E-02(-)&\textbf{8.16E-03}($\approx$)&5.23E-02(-)&6.26E-02(-)&3.91E-02(-)\\
                             &T2&1.87E-02&4.14E-01(-)&3.70E-01(-)&\textbf{1.37E-02}(+)&1.66E-01(-)&1.33E-01(-)&4.08E-01(-)\\
\hline

\multirow{2}{*}{CMO6}&T1&\textbf{7.73E-03}&2.95E-02(-)&2.97E-02(-)&1.36E-02(-)&4.91E-02(-)&6.71E-02(-)&4.45E-02(-)\\
                             &T2&\textbf{1.05E-02}&8.71E-02(-)&6.91E-02(-)&3.61E-02(-)&1.30E-01(-)&1.26E-01(-)&5.33E-02(-)\\
\hline

\multirow{2}{*}{CMO7}&T1&\textbf{8.06E-03}&3.02E-02(-)&3.00E-02(-)&8.27E-03(-)&3.94E-02(-)&4.52E-02(-)&2.36E-02(-)\\
                             &T2&\textbf{7.92E-03}&3.02E-02(-)&2.96E-02(-)&1.08E-02(-)&3.98E-02(-)&4.40E-02(-)&2.89E-02(-)\\
\hline

\multirow{2}{*}{CMO8}&T1&\textbf{8.95E-03}&3.69E-02(-)&3.39E-02(-)&1.22E-02(-)&4.00E-02(-)&4.85E-02(-)&2.89E-02(-)\\
                             &T2&\textbf{1.64E-02}&1.55E-01(-)&1.54E-01(-)&2.62E-02(-)&3.82E-01(-)&4.46E-01(-)&2.62E-01(-)\\
\hline

\multirow{2}{*}{CMO9}&T1&\textbf{1.84E-02}&3.59E-01(-)&3.58E-01(-)&3.16E-02(-)&2.93E-01(-)&2.57E-01(-)&7.01E-01(-)\\
                             &T2&\textbf{1.01E-02}&9.71E-02(-)&9.42E-02(-)&4.60E-02(-)&1.55E-01(-)&1.37E-01(-)&5.31E-02(-)\\
\hline

\multirow{2}{*}{CMO10}&T1&\textbf{1.72E-02}&1.51E-01(-)&1.32E-01(-)&5.95E-02(-)&3.91E-01(-)&4.46E-01(-)&2.42E-01(-)\\
                              &T2&\textbf{1.72E-02}&1.53E-01(-)&1.37E-01(-)&9.67E-02(-)&3.08E-01(-)&3.79E-01(-)&1.66E-01(-)\\
\hline

\multirow{1}{*}{$+/-/\approx$}&&&0/20/0&0/20/0&3/16/1&0/20/0&0/20/0&0/20/0\\
\bottomrule
\end{tabular}}
\end{table*}
EMT algorithms can work well on high similarity problems CI+HS, PI+HS and NI+HS. The overall performance of EMT algorithms is better than single-tasking algorithms. These results confirm that knowledge transfer across tasks in EMT is capable of accelerating convergence and finding better solutions.

For medium similarity problems CI+MS, PI+MS and NI+MS, EMT algorithms except for EMT-PD are not very competitive. CI+MS, PI+MS and NI+MS are comprised of unimodal and multimodal functions. The unimodal function only has a single global optimal solution. With the development of evolution, the population of unimodal function will gradually converge to a small area of search space. On the contrary, multimodal function has multiple global optimal solutions or local optimal solutions, the population of it will gradually converge to multiple different areas of search space. Since populations' convergence areas of unimodal and multimodal functions are quite different, the probability of negative transfers is high. MO-MFEA, TMO-MFEA and EMT-EGT cannot effectively handle those kinds of negative transfer, which makes their performance degradation. However, EMT-PD can reduce the probability of generating negative transfer and increase the diversity of population through the two-stage adaptive knowledge transfer, which makes a good performance of EMT-PD on medium similarity problems.

EMT algorithms are still competitive on CI+LS and PI+LS, although they belong to low similarity problems. Because the global optima of two tasks on CI+LS and PI+LS have identical variables in the unified search space, those two tasks can still provide useful knowledge to each other on identical variables. On the contrary, for NI+LS, all global optima variables of two tasks are different. These tasks can only provide less useful knowledge to each other, resulting in the poor performances of EMT algorithms proposed previously. EMT-PD still shows obvious competitiveness on NI+LS owing to the adaptively adjusted step size of search at the first stage of knowledge transfer.

\begin{figure}
\center
\fontsize{8}{9}\selectfont
\captionsetup{font={small}}
\includegraphics[width=8cm,height=5cm]{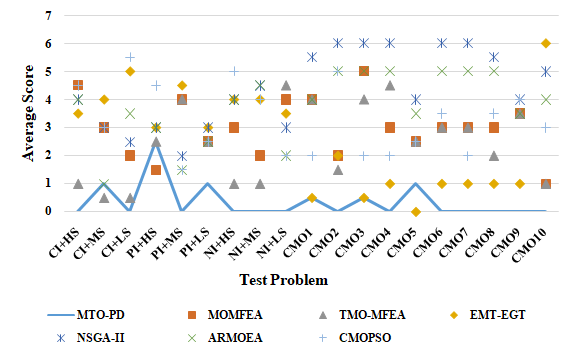}
\caption{The average performance score of all algorithms on MTMOPs and CEC2019-CMO. The values of EMT-PD are connected by a solid line to assess the score more easily.}\label{F5}
\end{figure}

\subsubsection {Results and Analysis on CEC2019-CMO}
\textbf{Table \ref{T3}} shows the experimental results on CEC2019-CMO. The overall performance of EMT-PD is better than other algorithms. CMO2, CMO4 and CMO10 are composed of functions with the same PS. Therefore, the diversity and convergence of population can be maintained well via knowledge transfer between tasks. Compared with single-task algorithms, the performance of EMT algorithms is excellent. For CMO1, CMO3 and CMO6, objective functions of two tasks are very similar, which means the two tasks have similar properties. EMT algorithms can effectively optimize this kind of problems. For CMO7, CMO8 and CMO9 with different PS and objective functions, the performances of EMT algorithms proposed previously are not significant. However, based on adaptive step size of searching at the first stage of knowledge transfer, EMT-PD can effectively reduce the probability of negative transfer. The PFs of two tasks in CMO5 are very complex. To achieve better performance, the diversity of population needs to be carefully maintained in the optimization process. EMT-EGT provides multiple searching mechanisms for the population, which is conducive to maintain the diversity of population and hence obtain best performance on CMO5.

Fig. \ref{F5} shows the average performance score of all algorithms on MTMOPs and CEC2019-CMO. The calculation method of average performance score can be found in \cite{Average_performance_score}. The lower average score is, the better performance of the IGD algorithm gets. EMT-PD has achieved the best results on MTMOPs and CEC2019-CMO, which demonstrates that the proposed knowledge extract and transfer method can effectively improve the performance of algorithm.

\subsection{Experiment and Analysis on Many-objective Problems}

\subsubsection {Test Problems and Compared Algorithms}
In order to further validate the performance of EMT-PD on MaOPs, we design a novel multi-tasking many-objective test suite based on the MaF test suite \cite{MaF}, labelled as MTMaOPs. The proposed MTMaOPs includes six problems, as shown in \textbf{Table \ref{T4}}. MaF-HS1 is composed of MaF3 and MaF4. MaF-HS2 is composed of MaF4 and MaF6. MaF-MS1 is composed of MaF1 and MaF5*, where MaF5* is a shifted MaF5. The shifted individual is $z=(p-\bm{r})$, where $\bm{r}=(0.05)^D$ and the original individual $p \in (0,1)^D$. MaF-MS2 is composed of MaF5 and MaF6*, where MaF6* uses the same shift operation as MaF5*. MaF-LS1 is composed of MaF4 and MaF5. MaF-LS2 is composed of MaF3 and MaF6. The similarity denoted by $sim$ of the problem is calculated as follows\cite{MTMOPs}:

\begin{equation}\label{sim}
sim = \frac{cov(\bm{R}_{1},\bm{R}_{2})}{std(\bm{R}_{1})std(\bm{R}_{2})},
\end{equation}
where $\bm{R}_{1}=(R_{1,1}, R_{1,2}, ..., ,R_{1,k},..., R_{1,K})$ and $\bm{R}_{2}=(R_{2,1}, R_{2,2}, ..., ,R_{2,k},..., R_{2,K})$, $k = (1, 2,..., K)$, $R_{1,k}$ and $R_{2,k}$ is the rank of the $k$-$th$ solution in population with respect to the two tasks. To calculate the value of $sim$, 1,000,000 solutions are randomly generated, that’s means $K$ =1,000,000.

\begin{table}[htbp]
  \centering
  \linespread{1.2}
  \captionsetup{font={small}}
  \setlength{\tabcolsep}{2mm}{
  \fontsize{8}{9}\selectfont
  \begin{threeparttable}
  \caption{SUMMARY OF THE PROPOSED MTMaOPs}\label{T4}
  \label{mut2}
    \begin{tabular}[c]{cccm{1cm}<{\centering}m{1cm}<{\centering}m{1cm}<{\centering}}
    \toprule
    Problem &Task&Function&$Rs$ with $m = 10$&$Rs$ with $m=20$&$Rs$ with $m=30$\cr
    \midrule
    \renewcommand{\arraystretch}{2}
    \multirow{2}{*}{MaF-HS1}&T1&MaF3&\multirow{2}{*}{1}&\multirow{2}{*}{1}&\multirow{2}{*}{1}\\&T2&MaF4\\
    \hline
    \multirow{2}{*}{MaF-HS2}&T1&MaF4&\multirow{2}{*}{1}&\multirow{2}{*}{1}&\multirow{2}{*}{1}\\&T2&MaF6\\
    \hline
    \multirow{2}{*}{MaF-MS1}&T1&MaF1&\multirow{2}{*}{0.3703}&\multirow{2}{*}{0.3756}&\multirow{2}{*}{0.4236}\\&T2&MaF5*\\
    \hline
    \multirow{2}{*}{MaF-MS2}&T1&MaF5&\multirow{2}{*}{0.3865}&\multirow{2}{*}{0.3866}&\multirow{2}{*}{0.4396}\\&T2&MaF6*\\
    \hline
    \multirow{2}{*}{MaF-LS1}&T1&MaF4&\multirow{2}{*}{0.0038}&\multirow{2}{*}{0.0051}&\multirow{2}{*}{0.0044}\\&T2&MaF5\\
    \hline
    \multirow{2}{*}{MaF-LS2}&T1&MaF3&\multirow{2}{*}{0.0038}&\multirow{2}{*}{0.0052}&\multirow{2}{*}{0.0051}\\&T2&MaF6\\
    \bottomrule
    \end{tabular}
    \end{threeparttable}
    }
\end{table}

The $sim$ lying in $(0,1/3]$, $(1/3,2/3]$, $(2/3,1]$ is regarded as low, medium, and high similarity, respectively. According to the degree of similarity, MTMaOPs is divided into high similarity problems: MaF-HS1 and MaF-HS2, medium similarity problems MaF-MS1 and MaF-MS2, low similarity problems MaF-LS1 and MaF-LS2. In addition, each problem of MTMaOPs is set to contain three instances of different objective dimensions, that is, $n=10$, $n=20$, and $n=30$.

Six state-of-the-art algorithms are used in comparision with EMT-PD on MTMaOPs, including three EMT algorithms, i.e., MO-MFEA \cite{MOMFEA}, TMO-MFEA \cite{TMO-MFEA}, EMT-EGT \cite{EMT-EGT}, and three many-objective optimization algorithms, i.e., NSGA-III \cite{NSGA-III}, VaEA \cite{VaEA} and DDEANS \cite{DDEANS}. NSGA-III is a basic many-objective evolutionary algorithms serving as the baseline here. VaEA is characterized by novel selection strategies. DDEANS is characterized by novel dynamical decomposition strategy.

\subsubsection {Performance Metric}
The Modified Inverted Generational Distance (IGD$_{+}$) \cite{IGDPLUS} is adopted in this paper as the performance evaluation measure on MTMaOPs. IGD$_{+}$ considers the Pareto dominance relation between reference vector and solution. It can accurately evaluate the performance of many-objective optimization algorithms. IGD$_{+}$ is calculated as follows:

\begin{equation}
IGD_{+}({{P}^{*}}, \bm{A})=\frac{1}{\left| {{P}^{*}} \right|}\sum\limits_{\bm{z}\in {{P}^{*}}}{min\sqrt{\sum_{k=1}^{n}(max(\{z_{k}-A_{k}\},0))^{2}}},
\end{equation}
where $\bm{z} = (z_{1}, z_{2}, ..., z_{n})$ represents a reference vector. $\bm{A} = (A_{1}, A_{2}, ..., A_{n})$ represents a solution set, $n$ is the dimension of objectives, and $P^*$ denotes a predefined set of reference points on the PF.  In this experiment, the number of reference points for calculating IGD$_{+}$ is set to 100,000. The smaller value of IGD$_{+}$ is, the better convergence and diversity of population gets.

\begin{table}[htbp]
\centering
\linespread{1.4}
\captionsetup{font={small}}
\caption{PARAMETERS SETTING FOR DIFFERENT OBJECTIVE DIMENSION ON MTaMOPs}\label{T5}
\setlength{\tabcolsep}{3mm}{
\fontsize{8}{9}\selectfont
\begin{tabular}{ccccc}
\toprule
$n$&$H$&$N$&$G$&FEs\\
\midrule
10&(3,1)&230&300&69000\\
20&(2,2)&420&300&126000\\
30&(2,1)&465&300&139500\\
\bottomrule
\end{tabular}
}
\end{table}

\begin{table}[htbp]
  \centering
  \linespread{1.2}
  \captionsetup{font={small}}
  \setlength{\tabcolsep}{4mm}{
  \fontsize{8}{9}\selectfont
  \begin{threeparttable}
  \caption{PARAMETER SETTINGS FOR CROSSOVER AND MUTATION ON MTaMOPs}
  \label{T6}
    \begin{tabular}{cc}
    \toprule
    Parameter&Value\cr
    \midrule
    \renewcommand{\arraystretch}{2}
    Crossover probability ($p_{c}$)&0.3\\
    Mutation probability ($p_{m}$)&$1 \slash N$\\
    Distribution index for crossover ($\eta_c$)&20\\
    Distribution index for mutation ($\eta_m$)&20\\
    \bottomrule
    \end{tabular}
    \end{threeparttable}
    }
\end{table}

\subsubsection {Parameter Settings}
Because decomposition-based algorithms NSGA-III and VaEA are limited by reference vector, the population size of NSGA-III and VaEA can not be set optionally. For fair comparison, the same setting of population size are employed for all algorithms. The setting of population size can refer to \cite{MOEADD}. The details of parameters for different objective dimension are summarized in \textbf{Table \ref{T5}}, where $n$ indicates the number of objectives, $H$ is the number of divisions considered along each objective coordinate, $N$ represent the size of population, $G$ is the maximal iteration, and FEs is maximal function evaluations. The common parameters of crossover and mutation are summarized in \textbf{Table \ref{T6}}.

\begin{table*}[tp]
\centering
\linespread{1}
\captionsetup{font={small}}
\fontsize{8}{9}\selectfont
\caption{AVERAGED IGD+ VALUE OBTAINED BY EMT-PD, MO-MFEA, TMO-MFEA, EMT-EGT, NSGA-III, VaEA AND DDEANS OVER 30 INDEPENDENT RUNS ON MTMaOPs}\label{T7}
\setlength{\tabcolsep}{2mm}{
\begin{tabular}{cccccccccc}
\toprule
Problem&Number of Objectives&Task&EMT-PD&MO-MFEA&TMO-MFEA&EMT-EGT&NSGA-III&VaEA&DDEANS\\
\toprule
\multirow{6}{*}{MaF-HS1}&\multirow{2}{*}{10}&T1&\textbf{2.03E-01}&5.39E+05(-)&5.48E+05(-)&6.44E+05(-)&1.74E+06(-)&4.24E+04(-)&1.26E+03(-)\\
                                           &&T2&\textbf{5.57E-01}&2.08E+01(-)&2.81E+01(-)&8.83E+01(-)&7.52E+01(-)&2.16E+01(-)&3.08E+01(-)\\
\cmidrule(lr){2-10}
&\multirow{2}{*}{20}&T1&\textbf{2.10E-01}&4.57E+02(-)&8.61E+02(-)&1.71E+02(-)&4.95E+03(-)&4.10E+04(-)&9.26E+02(-)\\
                   &&T2&\textbf{8.13E-01}&1.98E+03(-)&8.86E+03(-)&1.23E+03(-)&3.65E+04(-)&7.98E+03(-)&1.31E+04(-)\\
\cmidrule(lr){2-10}
&\multirow{2}{*}{30}&T1&\textbf{2.25E-01}&4.84E+03(-)&8.98E+03(-)&1.84E+03(-)&5.22E+03(-)&4.55E+04(-)&9.35E+02(-)\\
                   &&T2&\textbf{9.14E-01}&1.98E+03(-)&8.86E+03(-)&1.23E+03(-)&3.65E+04(-)&7.98E+03(-)&1.31E+04(-)\\
\cmidrule(lr){1-10}

\multirow{6}{*}{MaF-HS2}&\multirow{2}{*}{10}&T1&4.36E-01&2.00E+00(-)&\textbf{1.43E-01}(+)&2.06E-01(+)&4.04E+01(-)&2.74E+03(-)&9.53E+02(-)\\
                                           &&T2&\textbf{6.71E-01}&1.92E+00(-)&8.95E-01(-)&8.14E-01(-)&2.20E+00(-)&6.76E+00(-)&2.44E+00(-)\\
\cmidrule(lr){2-10}
&\multirow{2}{*}{20}&T1&4.85E-01&1.14E+00(-)&4.05E-01(+)&\textbf{3.11E-01}(+)&3.65E+04(-)&7.98E+03(-)&1.31E+04(-)\\
                   &&T2&\textbf{6.83E-01}&2.16E+01(-)&1.30E-01(-)&7.51E-01(-)&1.99E+00(-)&4.34E-01(-)&3.40E+00(-)\\
\cmidrule(lr){2-10}
&\multirow{2}{*}{30}&T1&5.51E-01&1.20E+00(-)&3.87E-01(+)&\textbf{3.22E-01}(+)&2.93E+07(-)&7.30E+06(-)&1.15E+07(-)\\
                   &&T2&\textbf{6.88E-01}&1.92E+01(-)&7.69E-01(-)&8.14E-01(-)&1.70E+00(-)&1.47E+00(-)&6.82E+00(-)\\
\cmidrule(lr){1-10}

\multirow{6}{*}{MaF-MS1}&\multirow{2}{*}{10}&T1&6.12E-01&8.86E-01(-)&2.47E-01(+)&2.06E-01(+)&2.16E-01(+)&\textbf{1.67E-01}(+)&1.70E-01(+)\\
                                           &&T2&\textbf{6.24E-01}&4.12E+00(-)&2.12E+00(-)&4.38E+00(-)&1.06E+00(-)&1.43E+00(-)&1.08E+00(-)\\
\cmidrule(lr){2-10}
&\multirow{2}{*}{20}&T1&4.90E-01&1.16E+00(-)&4.23E-01(+)&3.11E-01(+)&3.77E-01(+)&2.54E-01(+)&\textbf{2.51E-01}(+)\\
                   &&T2&\textbf{6.02E-01}&3.41E+00(-)&3.31E+00(-)&4.85E+00(-)&1.55E+00(-)&1.80E+00(-)&1.51E+00(-)\\
\cmidrule(lr){2-10}
&\multirow{2}{*}{30}&T1&6.41E-01&1.24E+00(-)&3.99E-01(+)&3.22E-01(+)&3.56E-01(+)&2.49E-01(+)&\textbf{2.48E-01}(+)\\
                   &&T2&\textbf{7.88E-01}&3.53E+00(-)&3.84E+00(-)&4.95E+00(-)&1.69E+00(-)&1.89E+00(-)&1.69E+00(-)\\
\cmidrule(lr){1-10}

\multirow{6}{*}{MaF-MS2}&\multirow{2}{*}{10}&T1&\textbf{6.09E-01}&1.07E+00(-)&4.12E+00(-)&4.11E+00(-)&1.08E+02(-)&1.06E+00(-)&1.43E+00(-)\\
                                          &&T2&\textbf{4.00E-03}&2.36E+01(-)&2.50E+00(-)&3.27E-01(-)&1.31E+00(-)&9.89E-01(-)&1.44E+00(-)\\
\cmidrule(lr){2-10}
&\multirow{2}{*}{20}&T1&\textbf{7.62E-01}&3.53E+00(-)&4.90E+00(-)&4.12E+00(-)&1.55E+00(-)&1.79E+00(-)&1.51E+00(-)\\
                   &&T2&\textbf{4.65E-03}&2.06E+01(-)&5.80E+00(-)&6.20E-01(-)&1.98E+00(-)&4.35E-01(-)&3.40E+00(-)\\
\cmidrule(lr){2-10}
&\multirow{2}{*}{30}&T1&6.41E-01&3.65E+00(-)&4.96E+00(-)&4.96E+00(-)&1.69E+00(-)&1.89E+00(-)&1.68E+00(-)\\
                   &&T2&\textbf{7.88E-01}&2.01E+01(-)&5.95E+00(-)&8.17E-01(-)&1.70E+00(-)&1.46E+00(-)&6.82E+00(-)\\
\cmidrule(lr){1-10}

\multirow{6}{*}{MaF-LS1}&\multirow{2}{*}{10}&T1&\textbf{6.11E-01}&2.23E+05(-)&9.09E+03(-)&2.00E+04(-)&7.52E+01(-)&2.16E+01(-)&3.08E+01(-)\\
                                           &&T2&\textbf{6.22E-01}&4.14E+00(-)&2.22E+00(-)&8.64E+00(-)&1.06E+00(-)&1.43E+00(-)&1.08E+00(-)\\
\cmidrule(lr){2-10}
&\multirow{2}{*}{20}&T1&\textbf{1.69E+00}&2.02E+05(-)&7.72E+04(-)&3.00E+04(-)&3.65E+04(-)&7.98E+03(-)&1.31E+04(-)\\
                   &&T2&\textbf{5.04E-01}&3.77E+00(-)&3.10E+00(-)&1.03E+01(-)&1.55E+00(-)&1.80E+00(-)&1.51E+00(-)\\
\cmidrule(lr){2-10}
&\multirow{2}{*}{30}&T1&\textbf{2.12E+00}&1.99E+06(-)&6.83E+04(-)&3.72E+05(-)&2.93E+06(-)&7.30E+05(+)&1.15E+06(-)\\
                   &&T2&\textbf{6.13E-01}&3.74E+00(-)&3.70E+00(-)&1.11E+01(-)&1.69E+00(-)&1.89E+00(-)&1.68E+00(-)\\
\cmidrule(lr){1-10}

\multirow{6}{*}{MaF-LS2}&\multirow{2}{*}{10}&T1&\textbf{1.95E-01}&4.92E+05(-)&8.90E+05(-)&7.28E+05(-)&1.74E+06(-)&4.24E+04(-)&1.26E+03(-)\\
                                           &&T2&\textbf{2.38E-03}&2.26E+01(-)&1.23E-01(-)&3.49E-01(-)&1.31E+00(-)&9.91E-01(-)&1.44E+00(-)\\
\cmidrule(lr){2-10}
&\multirow{2}{*}{20}&T1&\textbf{2.56E-01}&4.13E+05(-)&1.04E+06(-)&1.70E+06(-)&4.95E+03(-)&4.10E+04(-)&9.26E+02(-)\\
                   &&T2&\textbf{1.82E-03}&2.06E+01(-)&1.37E-01(-)&6.66E-01(-)&1.99E+00(-)&4.35E-01(-)&3.40E+00(-)\\
\cmidrule(lr){2-10}
&\multirow{2}{*}{30}&T1&\textbf{3.38E-01}&4.14E+05(-)&1.13E+06(-)&2.38E+06(-)&1.74E+02(-)&1.01E+04(-)&6.59E+03(-)\\
                   &&T2&\textbf{2.31E-03}&1.93E+01(-)&3.29E-01(-)&8.33E-01(-)&1.69E+00(-)&1.46E+00(-)&6.83E+00(-)\\
\cmidrule(lr){1-10}
\multicolumn{1}{c}{$+/-/\approx$}&&&&0/36/0&6/30/0&6/30/0&3/33/0&3/33/0&3/33/0\\
\bottomrule
\end{tabular}
}
\end{table*}

\subsubsection {Results and Analysis on MTMaOPs}
\textbf{Table \ref{T7}} lists the statistical results on MTMaOPs. Except for MaF-HS2 and MaF-MS1, EMT-PD performs better than other algorithms. This result shows that the knowledge extract and transfer of EMT-PD is efficient on many-objective optimization problems. The results of MaF-HS2 and MaF-MS1 are analyzed in details as the following.

MaF-HS2 is a high similarity problem and consists of MaF4 and MaF6. The property of high similarity can promote the cooperation between two tasks, and improve the diversity and convergence of population for EMT algorithms. It makes all EMT algorithms have obvious competitiveness on MaF-HS2 compared with single-tasking many-objective algorithms. MaF4 is a badly scaled many-objective function. The traditional non-dominated sorting employed by EMT-PD cannot normalize the value of objective function, which leads to the fact that  the population of MaF4 tends to converge to one side of PF, and cannot guide the search of population toward to convergence area effectively. It makes the poor performance of EMT-PD compared with EMT-EGT and TMO-MFEA.

MaF-MS1 is a medium similarity problem and is composed of MaF1 and MaF5$^*$. MaF1 is a linear problem, with its Pareto optimal solutions mainly concentrate on a very small area of searching space. In the early stage of evolution, only a few individuals can accurately reflect the centralization of population for MaF1. VaEA adopts maximum-vector-angle-first principle in environmental selection to ensure the diversity of population. DDEANS can balance dynamically the diversity and convergence of population according to the Euclidean distance between reference vector and population. The results on MaF-MS1 of VaEA and DDEANS are competitive just owing to the diversity mechanism of population effectively. TMO-MFEA uses different crossover parameters for different variables and EMT-EGT provides a variety of search mechanisms to the population. The diversity mechanism of TMO-MFEA and EMT-EGT are also superior to EMT-PD, which brings better performance of TMO-MFEA and EMT-EGT than EMT-PD on MaF-MS1.

\begin{figure}
\center
\fontsize{8}{9}\selectfont
\captionsetup{font={small}}
\includegraphics[width=7cm,height=4cm]{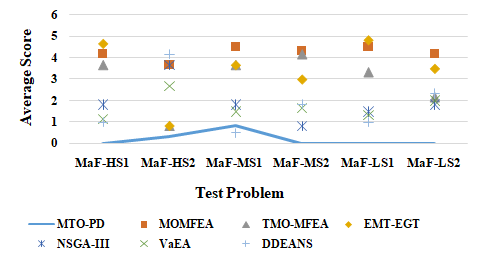}
\caption{The average performance score of all algorithms on MTMaOPs. The values of EMT-PD are connected by a solid line to assess the score more easily.}\label{F6}
\end{figure}

\begin{figure}
\center
\fontsize{8}{9}\selectfont
\captionsetup{font={small}}
\includegraphics[width=8cm,height=4.8cm]{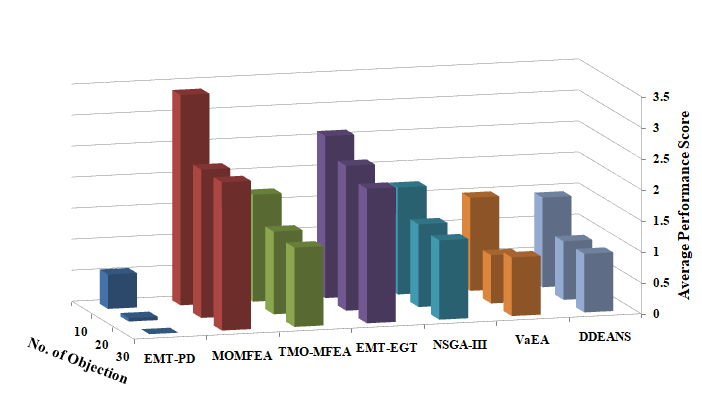}
\caption{The average performance score of all algorithms over 10, 20 and 30 objective dimensions of MTMaOPs.}\label{F7}
\end{figure}
Fig. \ref{F6} shows the average performance score of all algorithms on MTMaOPs. The smaller the score is, the better IGD$_{+}$ the algorithm gets. It can be seen that single-tasking many-objective optimization algorithms achieves better performance than EMT algorithms on most of the test problems, which illustrates that EMT algorithm proposed previously cannot effectively deal with MaOPs. EMT-PD is still competitive on MaOPs, because the search direction of each variable is guided by population distribution, which makes the knowledge transfer is more accurate.

Fig. \ref{F7} shows the average performance score of all algorithms over 10, 20 and 30 objective dimensions for MTMaOPs in term of IGD$_{+}$. It can be seen that with the increase of objective dimension, the average performance score of EMT-PD becomes lower, which indicates that EMT-PD has more advantages in handling optimization problems with larger objective dimensions.

\begin{figure}
\centering
\captionsetup{font={footnotesize}}
\subfigure[]{
\begin{minipage}{4cm}
\includegraphics[width=4.2cm,height=3cm]{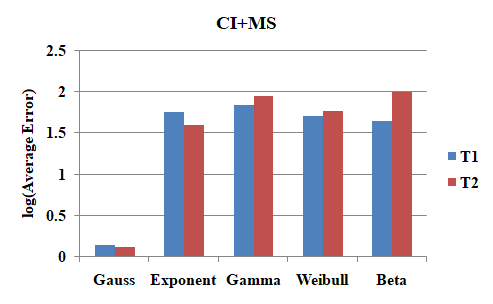}
\end{minipage}
}
\subfigure[]{
\begin{minipage}{4cm}
\centering
\includegraphics[width=4.2cm,height=3cm]{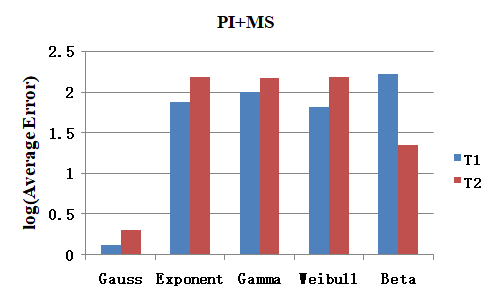}
\end{minipage}
}
\\
\subfigure[]{
\begin{minipage}{4cm}
\centering
\includegraphics[width=4.2cm,height=3cm]{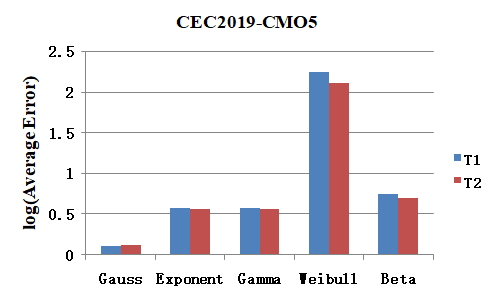}
\end{minipage}
}
\subfigure[]{
\begin{minipage}{4cm}
\centering
\includegraphics[width=4.2cm,height=3cm]{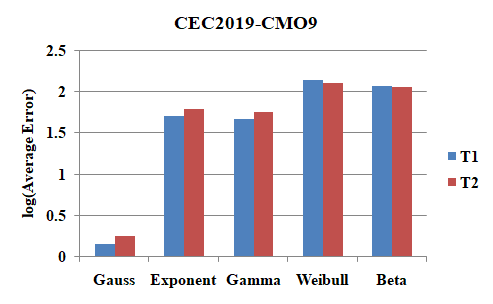}
\end{minipage}
}
\\
\subfigure[]{
\begin{minipage}{4cm}
\centering
\includegraphics[width=4.2cm,height=3cm]{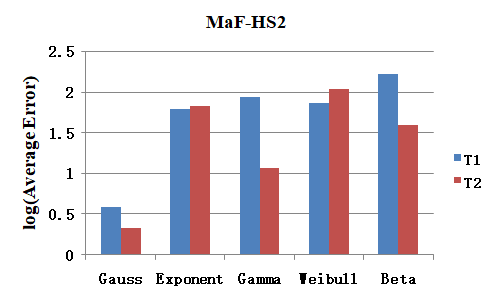}
\end{minipage}
}
\subfigure[]{
\begin{minipage}{4cm}
\centering
\includegraphics[width=4.2cm,height=3cm]{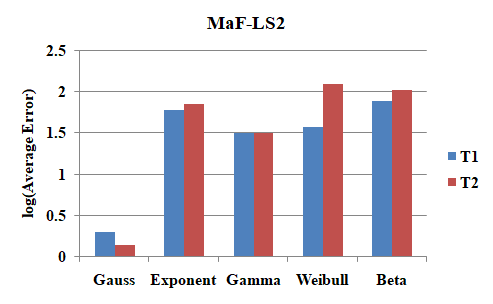}
\end{minipage}
}
\caption{Average fitting error of EMT-PD with different probability models on some problems.}\label{F8}
\end{figure}

\begin{figure*}
\center
\fontsize{8}{9}\selectfont
\captionsetup{font={small}}
\subfigure[]{
\begin{minipage}{4.2cm}
\includegraphics[width=4.2cm,height=3cm]{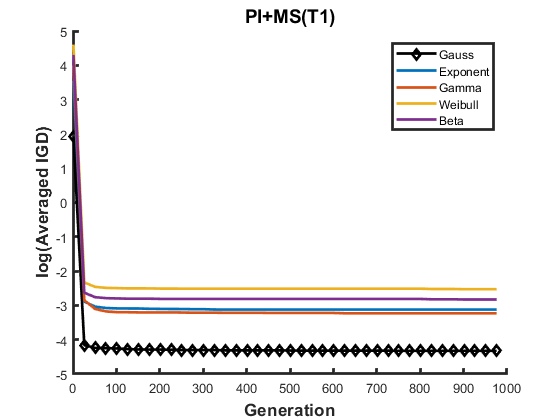}
\end{minipage}
\label{F9a}
}
\subfigure[]{
\begin{minipage}{4.2cm}
\centering
\includegraphics[width=4.2cm,height=3cm]{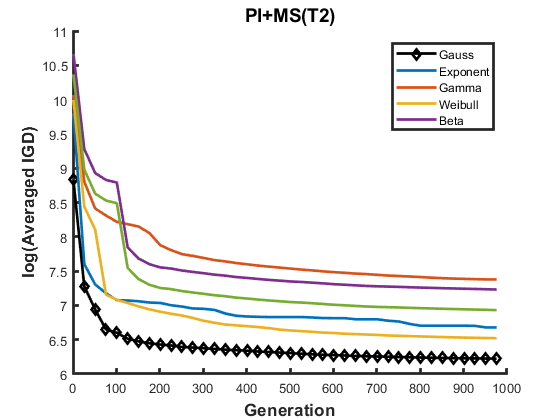}
\end{minipage}
\label{F9b}
}
\subfigure[]{
\begin{minipage}{4.2cm}
\centering
\includegraphics[width=4.2cm,height=3cm]{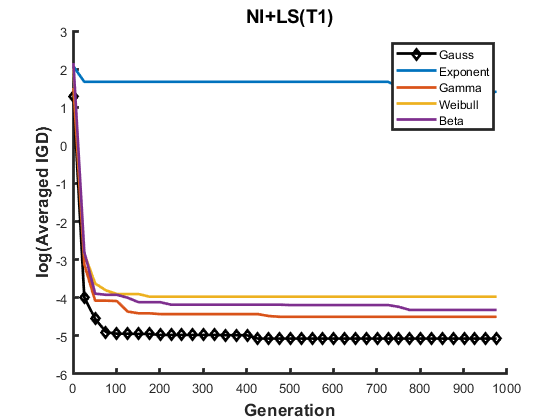}
\end{minipage}
\label{F9c}
}
\subfigure[]{
\begin{minipage}{4.2cm}
\centering
\includegraphics[width=4.2cm,height=3cm]{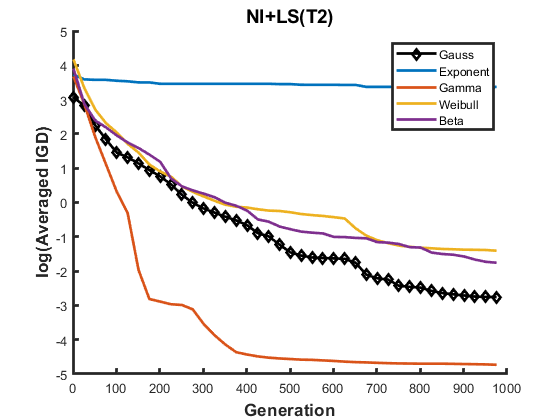}
\end{minipage}
\label{F9d}
}
\\
\subfigure[]{
\begin{minipage}{4.2cm}
\includegraphics[width=4.2cm,height=3cm]{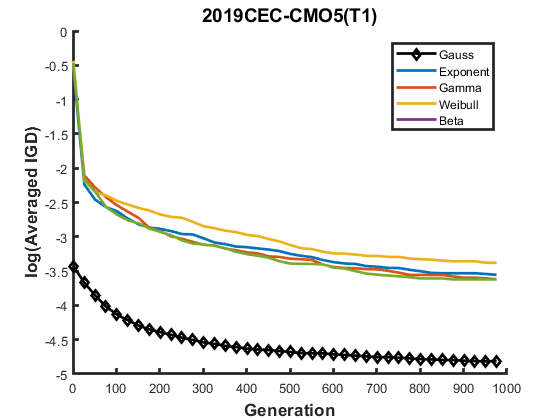}
\end{minipage}
\label{F9e}
}
\subfigure[]{
\begin{minipage}{4.2cm}
\centering
\includegraphics[width=4.2cm,height=3cm]{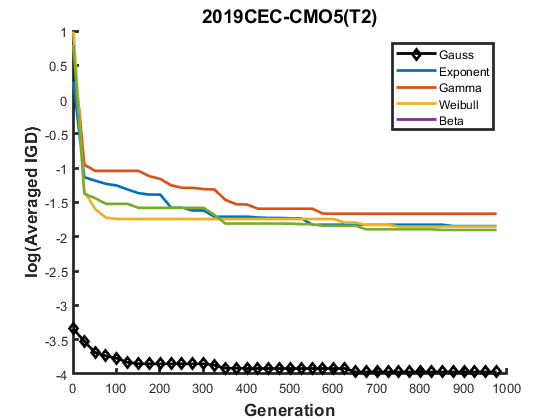}
\end{minipage}
\label{F9f}
}
\subfigure[]{
\begin{minipage}{4.2cm}
\centering
\includegraphics[width=4.2cm,height=3cm]{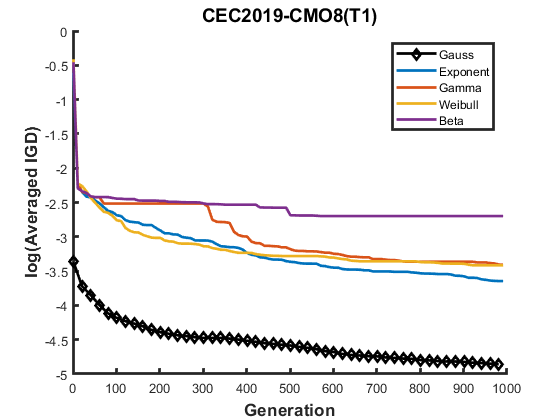}
\end{minipage}
\label{F9g}
}
\subfigure[]{
\begin{minipage}{4.2cm}
\centering
\includegraphics[width=4.2cm,height=3cm]{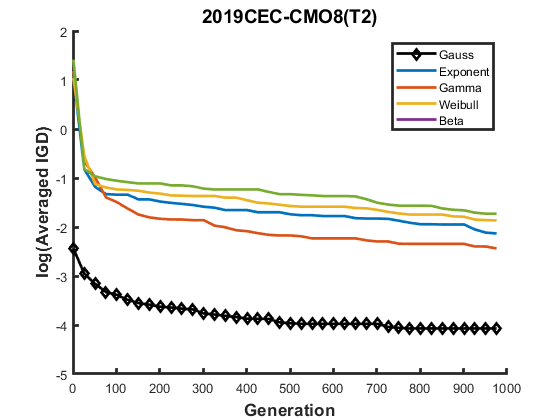}
\end{minipage}
\label{F9h}
}
\\
\subfigure[]{
\begin{minipage}{4.2cm}
\includegraphics[width=4.2cm,height=3cm]{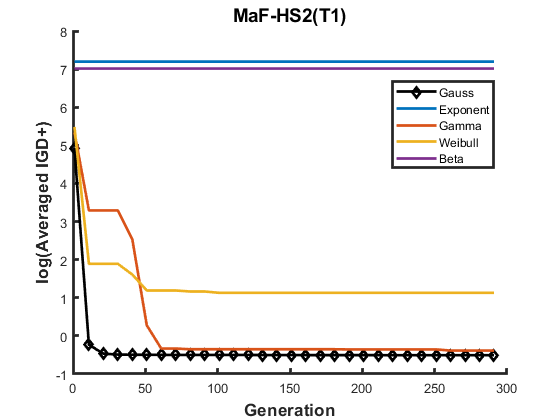}
\end{minipage}
\label{F9i}
}
\subfigure[]{
\begin{minipage}{4.2cm}
\includegraphics[width=4.2cm,height=3cm]{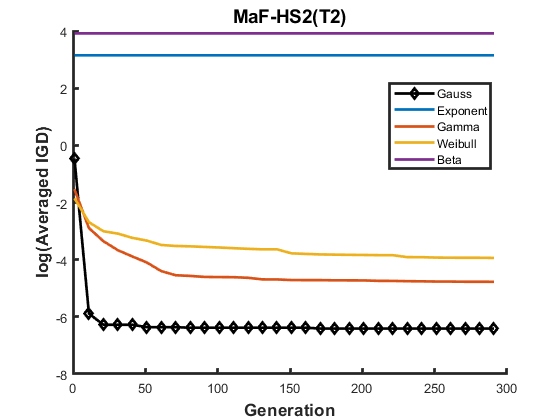}
\end{minipage}
\label{F9j}
}
\subfigure[]{
\begin{minipage}{4.2cm}
\includegraphics[width=4.2cm,height=3cm]{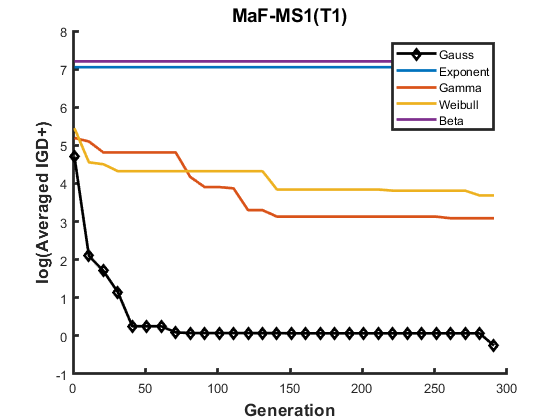}
\end{minipage}
\label{F9k}
}
\subfigure[]{
\begin{minipage}{4.2cm}
\includegraphics[width=4.2cm,height=3cm]{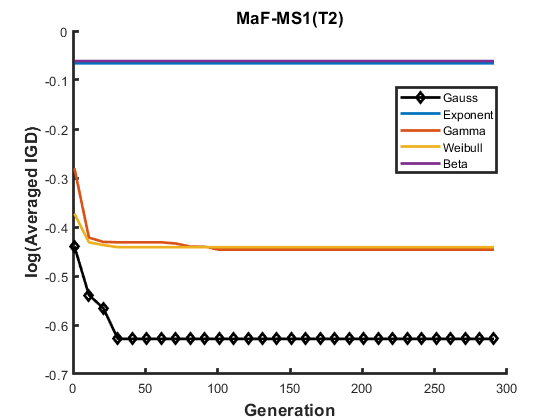}
\end{minipage}
\label{F9l}
}
\caption{IGD$\backslash$IGD$_{+}$ performance of EMT-PD with different probability models on some problems.}
\label{F9}
\end{figure*}

\section{THE PERFORMANCE OF EMT-PD WITH DIFFERENT PROBABILITY MODELS}
EMT-PD supports different probability models to fit population distribution. This section discusses the performance of EMT-PD with different probability models.

\subsection{The Error of Fitting Population Distribution with Different Probability Models}
Fig. \ref{F8} shows the average error of fitting population distribution with different probability models on CI+MS, PI+MS, 2019CEC-CMO5, 2019CEC-CMO9, MaF-HS2 and MaF-LS2. The error $e_{g}$ in each generation is calculated as follows\cite{MMSE}:

\begin{equation}\label{E12}
e_{g}=\frac{1}{N}\frac{1}{D}\sum_{i=1}^{N}\sum_{j=1}^{D}|1-M_{g,j}(x_{i,j})|,
\end{equation}
where $x_{i,j}$ represents the $j$-$th$ dimension of $i$-$th$ individuals, $M_{g,j}(x)$ represents the probability model of $j$-$th$ dimension calculated by Algorithm \ref{A2} at the $g$-$th$ generation. $g = 1,2,...,G$ and $G$ represents the maximum number of iterations. $N$ is the population size, and $D$ is the dimension of decision variable. The average error $e_{avg}$  in each generation is calculated as follows:

\begin{equation}\label{E14}
e_{avg}=\log(\frac{1}{G}\sum_{g=1}^{G}e_{g}),
\end{equation}

It can be seen from Fig. \ref{F8} that the average error of EMT-PD with Gaussian probability model is very small, especially for CI+MS, PI+MS, CEC2019-CMO9, MAF-HS2 and MAF-LS2. For CEC2019-CMO5, the average error of Exponential probability model, Gamma probability model and Beta probability model are also small. In summary, Gaussian probability model is the most versatile model for fitting all kinds of test suites, and other probability models may also be suitable for specific problems. In this paper, it is worth noting that the base of the logarithm in all experimental results is natural number $e$.

\subsection{IGD and IGD$_{+}$ of EMT-PD with Different Probability Models}
Fig. \ref{F9} (a)-(h) shows the average IGD of EMT-PD with different probability models on 30 independent running on PI+MS, NI+LS, 2019CEC-CMO5 and 2019CEC-CMO8. Fig. \ref{F9} (i)-(l) shows the average IGD$_{+}$ of EMT-PD with different probability models on 30 independent running on MAF-HS2 and MAF-MS1. It can be seen that EMT-PD with Gaussian probability model has better performance than EMT-PD with other probability models. EMT-PD with exponential model and Beta model shows poor performance on NI+LS, MAF-HS2 and MAF-MS1, which indicates that the universality of exponential model and Beta model are weak. It is important to note that the convergence speed of EMT-PD with Gamma model is faster than with Gaussian model on NI+LS in Fig. \ref{F9d}, which means that a strategy with adaptive selecting probability model is more suitable for the improvement of performance. The performance results on other problems with different probability models can refer to Section \uppercase\expandafter{\romannumeral5} of the \textbf{supplementary materials}.

\section{EFFECTS OF KNOWLEDGE TRANSFER FROM DIFFERENT INDIVIDUALS}
To echo the research motivations and further demonstrate the effectiveness of knowledge transfer from different individuals, this section discusses five variants of EMT-PD.

EMT-SR is the variant of EMT-PD, where the knowledge is randomly extracted from one individual of population. EMT-MR is the variant of EMT-PD, where the knowledge is randomly extracted from three individuals of population. EMT-SH is the variant of EMT-PD, where the knowledge is extracted from the optimal solution of population. EMT-MH is the variant of EMT-PD, where the knowledge is extracted from three optimal solutions of population. The optimal solution is the individual with the smallest Euclidean distance from PF. EMT-PD-1 is the variant of EMT-PD, which only has the first stage knowledge transfer. The comparative experiments are performed on five high similarity problems CI+HS, PI+HS, NI+HS, MaF-HS1, MaF-HS2, and five low similarity problems CI+LS, PI+LS, NI+LS, MaF-LS1, MaF-LS2. The experimental results can refer to Section \uppercase\expandafter{\romannumeral1} of the \textbf{supplementary materials}.

According to experimental results, it can be seen that the variants EMT-SR and EMT-MR extracting knowledge from one or more random individuals cannot converge effectively, and the variants EMT-SH and EMT-MH extracting knowledge from one or more optimal solutions cannot converge steadily. The performance of variant EMT-PD-1 is not good due to the worse diversity of population and the disadvantage of easily falling into local optimum. EMT-PD has the best performance than its variants on both high similarity problems and low similarity problems, which clarifies the motivation of this paper.

\section{CONCLUSION}
This paper proposes a new multi-objective EMT algorithm with two-stage adaptive knowledge transfer based on population distribution. The knowledge extracted from probability model can effectively guide the search of population for convergence. The first stage of knowledge transfer is characterized by a novel adaptive weight, which can effectively reduce the probability of generating negative transfer. At the second stage of knowledge transfer, the search range of individuals is adjusted dynamically again to balance the diversity and the convergence of population and to help jumping across local optimum. At the same time, in order to further study the performance of EMT-PD on MaOPs, a novel test suite MTMaOPs based on MaF test suite is proposed. EMT-PD is compared with state-of-the-art algorithms on MTMOPs, CEC2019-CMO and MTMaOPs. The experimental results show that EMT-PD is competitive.

Although EMT-PD has shown superior performance on various test suites, there are still some further works worth doing. For example, all kinds of real-world problems are very different and complex. Gaussian probability model may not be the best model to fit specific real-world problem. It is valuable to propose an adaptive selection strategy of probability model according to the characteristics of problem. In addition, many-tasking optimization is also an interesting field worth investigation. We will try to expand EMT-PD to many-tasking optimization in the future.

\bibliographystyle{IEEEtran}
\small
\bibliography{EMT-PD}
\end{document}